\documentclass{article}
\usepackage{ijcai25}

\usepackage{times}
\usepackage{soul}
\usepackage{url}
\usepackage[hidelinks]{hyperref}
\usepackage[utf8]{inputenc}
\usepackage[small]{caption}
\usepackage{graphicx}
\usepackage{amsmath,amssymb,amsfonts}
\usepackage{booktabs}
\usepackage{algorithm}
\usepackage{algorithmic}
\usepackage[switch]{lineno}

\usepackage{enumitem}
\usepackage{multirow}
\usepackage{adjustbox}
\usepackage{subcaption}
\usepackage{color}
\usepackage{amsmath,amssymb,amsfonts}
\usepackage{array}
\usepackage{booktabs}
\usepackage{adjustbox}
\usepackage[table]{xcolor}
\usepackage{url}
\usepackage{hyperref}
\usepackage{comment}

\title{What is Beneath Misogyny: Misogynous Memes Classification and Explanation}

\author{
	Kushal Kanwar\textsuperscript{\rm 1}
	\and
	Dushyant Singh Chauhan\textsuperscript{\rm 2}
	\and
	Gopendra Vikram Singh\textsuperscript{\rm 3}
	\and
	Asif Ekbal\textsuperscript{\rm 4}
	\affiliations
	{
		\textsuperscript{\rm 1}Jaypee University of Information Technology, Solan, India\\
		\textsuperscript{\rm 2}\href{https://cortextor.com/}{CortexTor Labs, India}\\
		\textsuperscript{\rm 3}Amity Center for Artificial Intelligence\\
		\textsuperscript{\rm 4}Indian Institute of Technology Jodhpur, India\\
	}
    \emails
    { \textnormal{ \{kushalneo, dushyantchauhan27, gopendra.99, asif.ekbal\}@gmail.com}\\}}

\begin{document}

\maketitle

\begin{abstract}
 Memes are popular in the modern world and are distributed primarily for entertainment. However, harmful ideologies such as misogyny can be propagated through innocent-looking memes. The detection and understanding of why a meme is misogynous is a research challenge due to its multimodal nature (image and text) and its nuanced manifestations across different societal contexts. We introduce a novel multimodal approach, \textit{namely}, \textit{\textbf{MM-Misogyny}} to detect, categorize, and explain misogynistic content in memes. \textit{\textbf{MM-Misogyny}} processes text and image modalities separately and unifies them into a multimodal context through a cross-attention mechanism. The resulting multimodal context is then easily processed for labeling, categorization, and explanation via a classifier and Large Language Model (LLM). The evaluation of the proposed model is performed on a newly curated dataset (\textit{\textbf{W}hat's \textbf{B}eneath \textbf{M}isogynous \textbf{S}tereotyping (WBMS)}) created by collecting misogynous memes from cyberspace and categorizing them into four categories, \textit{namely}, Kitchen, Leadership, Working, and Shopping. The model not only detects and classifies misogyny, but also provides a granular understanding of how misogyny operates in domains of life. The results demonstrate the superiority of our approach compared to existing methods. The code and dataset are available at \href{https://github.com/kushalkanwarNS/WhatisBeneathMisogyny/tree/main}{https://github.com/Misogyny}.   
\end{abstract}

\section{Introduction}
In today's era, social networks and online platforms are popular mediums for sharing thoughts and views in the form of micro blogs, images, stories, and memes \cite{joshi2024contextualizing}\footnote{\textbf{ChatGPT is used to refine and rephrase English content.}}. Touches of humor are often shared with docile memes, but dissemination of harmful ideologies and biases such as misogyny is also possible on these platforms \cite{rizzi2023recognizing}. The reinforcement of gender stereotypes promotes social inequalities and leads to harmful online and social environments. It is essential to detect and mitigate harmful content shared through memes. Since the number of users of on-line platforms is in the tens of millions, the only viable approach to address this issue is to automate the detection of misogynistic content and also comprehend its contextual manifestation in online spaces.

Traditional strategies to mitigate online misogyny, such as keyword filtering, rule-based methods, etc., are easily circumvented because they fail to capture subtleties inherent in multimodal misogynistic content \cite{hwang2023memecap}. Moreover, traditional methods offer no or limited interpretability, which makes it hard to understand the rationale behind recognizing certain content as misogynistic. It makes it difficult to raise awareness and educate users through a meaningful dialog about the detrimental effects of gender discrimination \cite{rizzi2023recognizing}. To overcome the limitations of traditional methods and address this complex issue, we propose a novel multimodal framework, \textit{namely} \textbf{\textit{MM-Misogyny}}, that detects, categorizes, and explains the misogyny in memes.

The main objective of this work is to explore how misogyny manifests itself in different spheres of life. For this work, misogyny is defined as any expression that includes undignified, hateful, mocking, or any gender stereotyping promoting bias against women. A harmless phrase and/or image that looks like it when used in a particular context can be highly offensive to women. The key aspect of our work is the recognition of misogyny in a contextual setting. Our framework categorizes misogyny into specific four domains, \textit{namely}, \textbf{Kitchen}, \textbf{Leadership}, \textbf{Working}, and \textbf{Shopping}. To account for the cases that fall outside these predefined contexts, the additional ``\textbf{Other}" category is assumed. This nuanced classification reveals misogyny's distribution across life aspects.

MM-Misogyny constitutes a multimodal architecture that exploits the text and image modalities of memes. The semantic features of text and image are extracted via separate encoders \textbf{Llama-3-8B} and \textbf{CLIP-ViT}, respectively. The features obtained from text and image modalities are then fused into a multimodal context by a cross-attention mechanism. This is subsequently processed by a classifier and a Large Language Model (LLM) to label memes misogynistic/non-misogynistic, categorize them into a particular class, and a detailed and transparent explanation for labeling a meme as misogynistic. 

The proposed approach goes beyond the simple detection of misogynistic content as it provides a comprehensive understanding of how misogyny operates in online spaces via memes. It also categorizes misogyny by social domains and enables the identification of areas in which gender bias is more prevalent. The explanation component of the proposed framework enhances the transparency of the detection process and allows users to understand the reasoning behind classification into misogynous categories. This leads to a more informed and constructive dialog with users on the harmful impact of misogynistic memes and the possibility of reinforcing positive behavior. 

The primary contributions of this work are as follows.
\begin{enumerate}[nolistsep]
	\item A novel multimodal framework, \textbf{\textit{MM-Misogyny}}, for the detection, categorization and explanation of misogyny in memes, designed to address the complexity of gender-based discrimination in online environments.
	\item The curation of new dataset, namely, \textit{\textbf{What's Beneath Misogynous Stereotyping (WBMS)}}. It follows a comprehensive categorization scheme that classifies misogyny within distinct social domains, including Kitchen, Leadership, Working, and Shopping.
	\item The innovative use of Large Language Models (LLMs) for both misogyny detection and the generation of transparent explanations, harnessing the advanced reasoning capabilities of these models to provide deeper insights.
	\item Extensive experiments conducted on \textbf{\textit{WBMS}} dataset demonstrate the proposed approach's effectiveness and accuracy in real-world scenarios.
\end{enumerate}

\paragraph{\textbf{Alignment with UN SDGs and the Leave No One Behind (LNOB) Principle.}}
This research contributes to several key United Nations Sustainable Development Goals (UN SDGs), particularly in the areas of gender equality and the reduction of social inequalities. By addressing the prevalent issue of misogyny in online content, we directly contribute to \textit{Goal 5: Gender equality}, working to challenge harmful stereotypes and promote a culture of respect and inclusion in digital spaces. Furthermore, by fostering a more inclusive online environment, we support \textit{Goal 10: Reduced Inequalities} and \textit{Goal 16: Peace, Justice, and Strong Institutions}, which aim to combat hate speech and promote tolerance in digital discourse. Our work is aligned with the Leave No One Behind (LNOB) principle, striving to create a safer, more equitable online experience for all individuals, regardless of gender, and contributing to the global effort to combat discrimination in all its forms.

\begin{figure*}[htbp]
	\centering
	\begin{subfigure}[b]{0.23\textwidth}
		\centering
		\includegraphics[height=3cm]{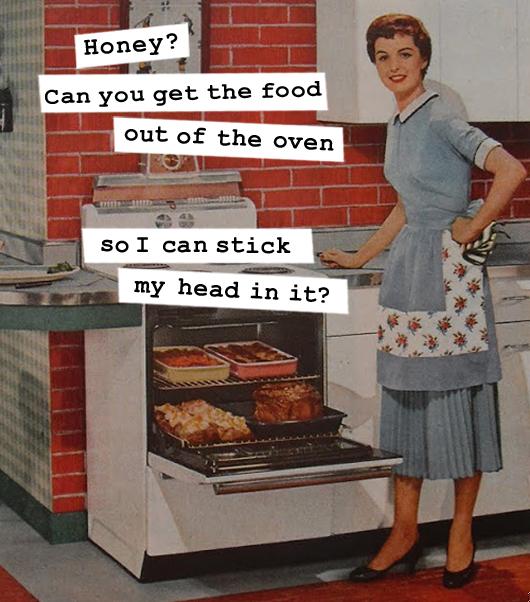}
		\caption{Kitchen}
		\label{fig:image1}
	\end{subfigure}
	\hfill
	\begin{subfigure}[b]{0.23\textwidth}
		\centering
		\includegraphics[height=3cm]{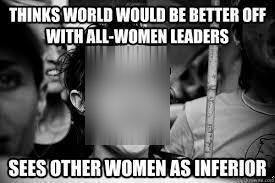}
		\caption{Leadership}
		\label{fig:image2}
	\end{subfigure}
	\hfill
	\begin{subfigure}[b]{0.23\textwidth}
		\centering
		\includegraphics[height=3cm]{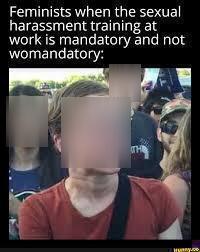}
		\caption{Working}
		\label{fig:image3}
	\end{subfigure}
	\hfill
	\begin{subfigure}[b]{0.23\textwidth}
		\centering
		\includegraphics[height=3cm]{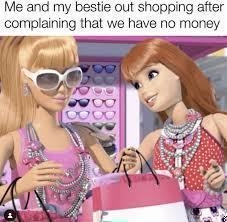}
		\caption{Shopping}
		\label{fig:image4}
	\end{subfigure}
	
	\caption{A sample of abusive memes from DACAM that trivialize or desensitize abuses against children. Human faces are blurred to prevent identity disclosure and potential harassment.}
	\label{fig_memes_sample}
\end{figure*}

\section{Related Work}
Addressing the rising prevalence of online misogyny is a critical challenge, and researchers have explored various methodologies to identify and mitigate such content. Existing studies primarily fall into two categories: unimodal and multimodal approaches. The unimodal approach focuses on textual content, leveraging linguistic features to detect offensive language. In contrast, the multimodal approach integrates multiple data modalities, such as images and their accompanying text, recognizing that harmful content extends beyond textual elements. This section reviews prior work in both unimodal and multimodal misogyny detection.

\subsection{Unimodal Misogyny Analysis}
Early research in misogyny detection predominantly relied on unimodal feature engineering techniques, including N-grams \cite{anzovino2018automatic}, Bag of Words \cite{pamungkas2018automatic}, Term Frequency-Inverse Document Frequency \cite{bakarov2018vector}, and word embeddings \cite{garcia2021detecting}. Additionally, handcrafted linguistic and syntactic features, such as part-of-speech tags, sentence structure, sentiment analysis, and hate lexicons, have been explored to enhance misogyny detection \cite{anzovino2018automatic,garcia2021detecting}.

The adoption of transformer-based models has significantly advanced unimodal misogyny detection. These models generate rich contextual embeddings, improving classification performance. Fine-tuning pre-trained transformers has yielded promising results, as demonstrated by \cite{attanasio2022benchmarking,calderon2023enhancing,muti2022misogyny}, who trained ten transformer models across multiple languages, including English, Spanish, and Italian.

Despite their effectiveness, unimodal approaches, particularly those limited to textual analysis, face inherent challenges in capturing nuanced contextual cues. As digital communication increasingly incorporates multimedia elements, purely unimodal methods often struggle to interpret misogynistic content embedded within images, videos, and other multimodal formats. 

\subsection{Multimodal Misogyny Analysis}
To overcome the limitations of unimodal methods, recent studies have shifted to multimodal approaches that integrate both textual and visual data. Advanced techniques, such as transformer-based architectures \cite{samghabadi2020aggression}, multimodal feature fusion \cite{rizzi2023recognizing}, and the use of large, diverse datasets \cite{fersini2022semeval}, have contributed to more sophisticated and context-sensitive misogyny detection frameworks. These approaches enhance the model's ability to analyze implicit and explicit misogynistic cues across different content formats, offering a more comprehensive solution to online misogyny detection.

Memes blend images and text, often conveying implicit messages that can be challenging to interpret. Misogynistic memes, in particular, can perpetuate harmful stereotypes and degrade women, affecting their emotional well-being. \cite{muti2022unibo,gu2022mmvae} focus on detecting misogyny in memes by combining textual and visual information. Their Multimodal BERT Transformer integrates BERT for text processing and CLIP for image encoding, demonstrating superior accuracy compared to unimodal baselines. Similarly, (\cite{gu2022mmvae}) proposed a Multimodal-Multitask Variational Autoencoder framework for misogyny detection in memes, utilizing BERT, ResNet-50, and CLIP for feature extraction, outperforming unimodal models.

In recent work, \cite{zhang2022srcb} investigated the interplay between vision and language using multimodal pretrained models. Their findings suggest that fine-tuned multimodal models, particularly CLIP, outperform unimodal methods due to extensive pretraining. 
While existing work focuses solely on detecting misogyny in memes, our work goes further by also generating explanations and analyzing the underlying contextual factors. To our knowledge, this is the first attempt to combine detection with explanation, offering a more comprehensive understanding of misogyny in memes.

\section{Dataset}\label{datalabel}
We introduce the first-of-its-kind, dataset of  \textit{\textbf{What's Beneath Misogynous Stereotyping (WBMS)}} created by collecting misogynous memes from cyberspace and categorizing them into four categories, \textit{namely}, \textbf{Kitchen,  Leadership, Working, and Shopping}. The data collection and annotation details are explained in the rest of the section.

\subsection{Data Collection}
\textit{\textbf{WBMS}} is created by collecting and categorizing (\textbf{Kitchen,  Leadership, Working, and Shopping}) by scraping web pages and some manual downloads. A meme combines an image and caption that can create funny, ironic, or harmful content. The presence or absence of text in memes and the relationship between captions and overlayed text can be categorized as follows: Images Only, Images with Overlayed Text, Images with Captions Only, Images with both Captions and Overlayed Text. Within a category (\textbf{Kitchen,  Leadership, Working, and Shopping}) we have categorized a meme as Image Only, Image and Caption Text Same, and Image and Cation Different and named them \textbf{Image}, \textbf{Same}, and \textbf{Different}, respectively. The dataset statistics are given in Table~\ref{tab_dataset}. Each row describes the category of the meme, counts in that category, memes with Different Text from Captions, Same Text as Captions, and Image-only memes mention in the form of triple \textit{(Different, Same, Image)}. The last column describes the proportion of each category of the memes to give the idea of the relative proportions of the memes. The total number of memes is 2130. A sample of each category of memes is given in Figure~\ref{fig_memes_sample}.

\begin{table}[ht]
	\centering
	\renewcommand{\arraystretch}{1}
	\setlength\tabcolsep{5pt}
		\resizebox{0.45\textwidth}{!}
		{
			\begin{tabular}{c|c|c|c}
				\hline
				\textbf{Category} & \textbf{Count} & \textbf{(Different, Same, Image)} & \textbf{Proportion}\\ \hline
				Kitchen & 1076 & (780, 125, 171) & 0.51\\ \hline
				Leadership & 534 & (262, 0, 272) & 0.25\\ \hline
				Working & 321 & (151, 0, 170) & 0.15\\ \hline
				Shopping & 199 & (118, 4, 77) & 0.09\\
				\hline
				\textbf{Total} & 2130 & (1311, 129, 690) & 1.0 \\
				\hline
			\end{tabular}
		}
		\caption{Statistics of \textbf{WBMS} dataset.}
		\label{tab_dataset}
	\end{table}

\subsection{Data Annotation}  
The annotation process for the \textbf{WBMS} dataset involved categorizing memes based on textual and visual elements. A team of three annotators—two linguists with Ph.Ds and one computer scientist with a Ph.D.—performed the labeling. The annotators were trained to understand meme structures, including how captions and overlayed text interact with images to convey humor, irony, or other contextual meanings.  

Each annotator independently worked on a mutually exclusive subset of memes to ensure diverse perspectives and minimize bias. The primary goal was to classify memes into one of the pre-defined categories: \textbf{Kitchen, Leadership, Working, and Shopping}, while further labeling them based on text presence and alignment as \textbf{Image, Same, or Different}.  

\subsubsection{Annotation Process}  

\begin{itemize}[nolistsep]  
	\item \textit{Identifying Textual-Visual Relationships:} Annotators determined whether a meme contained overlayed text, a caption, or both, and assessed whether the text aligned with the image’s meaning.  
	\item \textit{Categorizing Text Presence and Alignment:} Each meme was labeled as \textbf{Image Only}, \textbf{Same} (Image and Caption Text are the Same), or \textbf{Different} (Image and Caption Text are Different).  
	\item \textit{Ensuring Consistency Across Categories:} For each category (e.g., Kitchen, Leadership), memes were reviewed to ensure proper classification, avoiding mislabeling due to ambiguous or context-dependent humor.  
\end{itemize}  

\textbf{Conflict Resolution:}  
When annotators disagreed on meme categorization, a senior annotator reviewed the discrepancies and made a final decision based on predefined labeling guidelines.  

\subsubsection{Review and Consistency Check}  
To validate the annotation process, a peer-review evaluation was conducted using \textbf{accuracy (A) and consistency (C) metrics}:  

\begin{itemize}[nolistsep]  
	\item \textbf{Accuracy (A):} Rated on a scale of 1-5, where 5 indicates a perfectly labeled meme and 1 indicates misclassification.  
	\item \textbf{Consistency (C):} Rated on a scale of 1-5, ensuring that similar memes received similar labels across categories.  
\end{itemize}  

The evaluation scores for each annotator are summarized in Table~\ref{tab_annotation_quality}, demonstrating high accuracy and consistency.  
\begin{table}[ht]
	\centering
	\renewcommand{\arraystretch}{1.2}
	\setlength\tabcolsep{8pt}
	\resizebox{0.45\textwidth}{!}
	{
		\begin{tabular}{c|c|c|c}
			\hline
			\textbf{Annotator} & \textbf{Accuracy (A)} & \textbf{Consistency (C)} & \textbf{Kappa ($\kappa$)} \\ \hline
			Annotator 1 & 4.7 & 4.5 & 0.82 \\ \hline
			Annotator 2 & 4.6 & 4.6 & 0.79 \\ \hline
			Annotator 3 & 4.8 & 4.7 & 0.84 \\ \hline
			\textbf{Average} & \textbf{4.7} & \textbf{4.6} & \textbf{0.82} \\ \hline
		\end{tabular}
	}
	\caption{Evaluation scores for annotation quality, including Accuracy (A), Consistency (C), and Fleiss' Kappa ($\kappa$) for inter-annotator agreement. Scores are on a scale of 1-5, and $\kappa$ values indicate strong agreement among annotators.}
	\label{tab_annotation_quality}
\end{table}

\subsubsection{Challenges}  
Annotating memes introduced unique challenges:  

\begin{itemize}[nolistsep]  
	\item \textbf{Ambiguity in Humor and Irony:} Some memes required subjective interpretation, making it difficult to determine the intended meaning, particularly in cases of sarcasm or double meanings.  
	\item \textbf{Ensuring Objective Categorization:} Since meme humor is culturally and socially influenced, ensuring objective classification required predefined guidelines to minimize personal bias.  
\end{itemize}  

\section{Methodology}

\begin{figure}[!t]
	\centerline{\includegraphics[scale=.6]{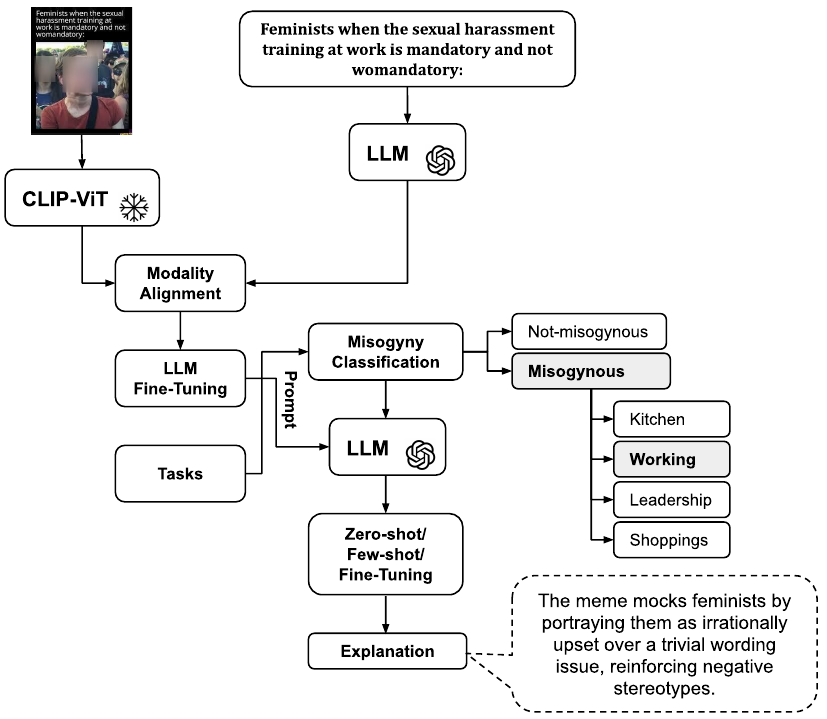}}
	\caption{Overall architecture for Multimodal Misogyny Analysis. \textbf{Please note that we concealed the person's face in the image, as demonstrated in the model, to uphold ethical standards and prevent potential harm. However, no such concealment was applied during training.}}\label{nsa_arch}
\end{figure}

In this section, we define the problem and describe our proposed framework, MM-Misogyny (Figure~\ref{nsa_arch}), designed for analyzing image and text modalities. Our primary objective is to classify whether a given meme exhibits misogyny, categorize the context of the misogyny, and generate an explanation for the detected misogyny.

\subsection{Problem Definition}

Each meme comprises a textual component \(T_k = (T_{k,1}, T_{k,2}, \ldots, T_{k,n})\), where \(n\) is the number of tokens, and an image component \(I_k\). These components are analyzed within the broader context \(H_k = ((T_1, I_1), (T_2, I_2), \ldots, (T_{k-1}, I_{k-1}))\). The framework aims to predict whether the \(k\)-th meme contains misogynistic content, designated as \(Y\), identify the context of the misogyny \(C \in \{\text{Kitchen, Leadership, Working, Shopping, Other}\}\), and provide reasoning for the detected misogyny.

\subsection{Multimodal Representation}

We employ separate encoders for textual and image modalities, aligning their representations within the text embedding space of the LLM.

\subsubsection{Textual Encoder}

We utilize the Llama-3-8B LLM \cite{touvron2023llama} as the textual encoder, leveraging its proficiency in language understanding and reasoning.

\subsubsection{Image Encoder}

The image modality is encoded using CLIP-ViT \cite{radford2021learning}, which effectively generates semantically rich representations by aligning images with textual descriptions. Specifically, we use the Vision Transformer (ViT)-based CLIP model to extract image embeddings.

\subsection{Modality Alignment}

Since LLMs primarily operate on textual inputs, the image embeddings are aligned to the LLM's textual embedding space for seamless integration. We employ a cross-attention mechanism to align image representations with the LLM's textual embeddings.

The attention mechanism is defined as:
\begin{equation}
	\text{Attention}(\mathbf{Q}, \mathbf{K}, \mathbf{V}) = \text{softmax}\left(\frac{\mathbf{Q}\mathbf{K}^T}{\sqrt{d_k}}\right) \mathbf{V}
\end{equation}

Let \(h_t\) and \(h_i\) represent the textual and image feature representations, respectively, where \(h_t \in \mathbb{R}^{L_t \times d_h}\) and \(h_i \in \mathbb{R}^{L_i \times d_h}\). A 1-D convolutional layer followed by a linear transformation is applied to the image features to align them with the LLM embedding space. The aligned representations are computed as:
\begin{equation}
	h'_i = \text{Linear}(\text{Conv1D}(h_i))
\end{equation}

The aligned image features \(h'_i\) are further refined using cross-attention:
\begin{equation}
	h^t_i = \text{Attn}(h'_i, E)
\end{equation}

Here, \(E\) denotes the embedding matrix associated with the LLM. The concatenation of aligned image and textual features constitutes the multimodal context:
\begin{equation}
	x = [h_t : h^t_i : \text{Embed}(\text{instruction})]
\end{equation}

\subsection{Misogyny Detection and Categorization}

The multimodal context \(x\) is passed to the LLM, which is prompted to classify whether the meme exhibits misogyny and categorize the context of the misogyny. The detection and categorization are performed in a zero-shot manner, leveraging the LLM's reasoning capabilities.  The LLM outputs both a binary misogyny classification (\(Y\)) and a category \(C \in \{\text{Kitchen, Leadership, Working, Shopping, Other}\}\).

\subsection{Explanation Generation}

Following the detection and categorization, a second LLM generates reasoning for the misogyny identified. The reasoning prompt includes:

\begin{enumerate}[nolistsep]
	\item The aligned multimodal context.
	\item The misogyny detection label \(\hat{y}\).
	\item The identified category \(\hat{C}\).
	\item Instructions to generate a detailed explanation for the detected misogyny, specifically relating it to the identified category.
\end{enumerate}

The reasoning LLM decomposes the task into subtasks, enabling accurate and interpretable results. This modular approach enhances performance, scalability, and interpretability while aligning with human cognitive processes.

\paragraph{\textbf{Inference:}}

In the inference step of the MM-Misogyny framework, we extract image and textual features using their respective encoders. These features are aligned via the cross-attention mechanism, producing multimodal representations that preserve the contextual integrity of the meme. The aligned features are used by the LLM to classify the presence of misogyny, identify its context, and generate reasoning. The equilibrium between textual and image modalities, along with fidelity to the original context, is maintained through the following formulation:
\begin{equation}
	\mathbf{z}_{test} = \mathcal{W}_{\alpha}(\mathcal{W}_{\omega}(h^t_i, h_i), \mathcal{W}_{\omega}(h_t, h^t_t))
\end{equation}

Here, $\mathcal{W}_{l}(c, d) = l \cdot c + (1 - l) \cdot d$, where \(\omega\) balances the contribution of image and text modalities. A higher \(\alpha\) indicates stronger fidelity to the original meme context. Customizing the trade-off between image and text fidelity is flexible based on meme characteristics and task-specific requirements. This flexibility allows users to optimize the balance between modalities for improved misogyny detection, categorization, and explanation.

\paragraph{\textbf{Component Design Justification:}}

The components of the MM-Misogyny framework are designed to address the multimodal nature of memes, ensuring accurate detection, categorization, and explanation of misogyny.
\begin{itemize}[nolistsep]
	\item Separate encoders for text (Llama-3-8B) and images (CLIP-ViT) ensure robust feature extraction, capturing distinct semantic characteristics from both modalities.
	\item The alignment process uses a cross-attention mechanism to map image features into the textual embedding space, facilitating seamless integration and reasoning by the LLM.
	\item The modular design of the framework ensures scalability, with the ability to handle complex multimodal tasks like aligning textual tone and visual content for accurate misogyny detection and categorization.
	\item The reasoning generation module uses a second LLM to provide interpretable and detailed explanations, ensuring transparency and trustworthiness in the detection and categorization process.
\end{itemize}

\begin{table*}[!ht]
	\centering
	\renewcommand{\arraystretch}{1}
	\setlength\tabcolsep{18pt}
	\begin{adjustbox}{max width=1\textwidth}
		\fontsize{10pt}{10pt}\selectfont
		\begin{tabular}{l|c|cccc}
			\hline
			\multirow{2}{*}{\bf Model} & \multirow{2}{*}{\bf MMC (F1)} & \multicolumn{4}{|c}{\textbf{Rationale Generation (RG)}} \\
			\cline{3-6}
			&  & \textbf{Relevance} & \textbf{Coherence} & \textbf{Readability} & \textbf{SemSim} \\ \hline
			\multicolumn{6}{c}{\textbf{Zero-shot Prompting for MMC and RG (Text-Only)}}\\ \hline
			Llama 3 & 0.76 & 0.78 & 0.79 & 0.76 & 0.78 \\
			Mistral 7B & 0.74 & 0.76 & 0.77 & 0.74 & 0.76 \\
			OpenHermes 2.5 & 0.73 & 0.75 & 0.76 & 0.73 & 0.75 \\
			\hline
			\multicolumn{6}{c}{\textbf{Few-shot Prompting for MMC and RG (Text-Only)}}\\ \hline
			Llama 3 & 0.79 & 0.81 & 0.80 & 0.79 & 0.80 \\
			Mistral 7B & 0.76 & 0.79 & 0.78 & 0.77 & 0.79 \\
			OpenHermes 2.5 & 0.75 & 0.77 & 0.76 & 0.75 & 0.78 \\
			\hline
			\multicolumn{6}{c}{\textbf{Fine-tuning (for MMC) + Prompting (for RG) (Text-Only)}}\\ \hline
			Llama 3 & 0.84 & 0.85 & 0.84 & 0.83 & 0.84 \\
			Mistral 7B & 0.82 & 0.83 & 0.82 & 0.81 & 0.83 \\
			OpenHermes 2.5 & 0.80 & 0.81 & 0.80 & 0.79 & 0.81 \\
			\hline 
			\rowcolor{green!20} \multicolumn{6}{c}{\textbf{\textit{[Ours]} Multimodal (Text + CLIP-ViT) Fine-tuning (for MMC) + Prompting (for RG)}}\\ \hline
			Llama 3 + CLIP-ViT & \bf 0.89 & \bf 0.89 & \bf 0.90 & \bf 0.91 & \bf 0.90 \\
			Mistral 7B + CLIP-ViT & \bf 0.86 & \bf 0.87 & \bf 0.88 & \bf 0.89 & \bf 0.89 \\
			OpenHermes 2.5 + CLIP-ViT & \bf 0.85 & \bf 0.86 & \bf 0.86 & \bf 0.87 & \bf 0.88 \\
			\hline
			\multicolumn{6}{c}{\textbf{\textit{[Ablation 1]} Multimodal Fine-tuning (for MMC) + Prompting (for RG) w/o CLIP-ViT}}\\ \hline
			Llama 3 & 0.85 & 0.86 & 0.85 & 0.86 & 0.87 \\
			Mistral 7B & 0.82 & 0.84 & 0.83 & 0.85 & 0.86 \\
			OpenHermes 2.5 & 0.81 & 0.83 & 0.82 & 0.84 & 0.85 \\
			\hline
			\multicolumn{6}{c}{\textbf{\textit{[Ablation 2]} Fine-tuning (for MMC) + Prompting (for RG) + w/ CLIP-ViT}}\\\hline
			Llama 3 + CLIP-ViT & 0.87 & 0.88 & 0.87 & 0.88 & 0.89 \\
			Mistral 7B + CLIP-ViT & 0.84 & 0.86 & 0.85 & 0.87 & 0.88 \\
			OpenHermes 2.5 + CLIP-ViT & 0.83 & 0.85 & 0.84 & 0.86 & 0.87 \\
			\hline
		\end{tabular}
	\end{adjustbox}
	\caption{Performance Comparison of Llama-3-8B, Mistral-7B, OpenHermes-2.5-Mistral  across different misogyny detection tasks in text-only and multimodal setups. MMC: Multimodal Misogyny Classification.}
	\label{tab:perf}
\end{table*}

\begin{table*}[h]
	\fontsize{10pt}{10pt}\selectfont
	\centering
	\begin{adjustbox}{max width=\textwidth}
		\begin{tabular}{|p{4cm}|p{4.2cm}|p{8cm}|}
			\hline
			\textbf{Post} & \textbf{Model Prediction} & \textbf{Rationale} \\ 
			\hline
			\small ``Women belong in the kitchen, and nowhere else. If she can’t cook, she isn’t wife material." & 
			\textbf{Llama 3-8B:}  \textcolor{blue}{Misogyny}  & The post explicitly promotes traditional gender roles and devalues women based on domestic abilities. The language used implies a rigid expectation that reinforces gender-based discrimination. \\ 
			\cline{2-3}
			& \textbf{Mistral 7B:} \textcolor{blue}{Misogyny}  & The post perpetuates harmful stereotypes about women's roles in society, associating their worth solely with domestic skills. The dismissive tone further supports the classification. \\ 
			\cline{2-3}
			& \textbf{OpenHermes 2.5-Mistral (w/ CLIP ViT):} \textcolor{blue}{Misogyny}  & The post conveys a direct misogynistic stereotype by implying that women are only valued for domestic abilities. The phrase “wife material” reinforces gendered expectations, making it a clear example of sexism. \\ 
			\hline
			\small ``My girlfriend tried to cook today… let’s just say we’re ordering takeout forever. Love her though!" & 
			\textbf{Llama 3-8B:} \textcolor{red}{Non-Misogyny} & While the post jokes about cooking ability, it does not reinforce traditional gender roles or degrade women. The inclusion of affectionate language suggests a lighthearted, non-misogynistic intent. \\ 
			\cline{2-3}
			& \textbf{Mistral 7B:}  \textcolor{red}{Non-Misogyny} & The post humorously discusses a personal anecdote without implying that a woman's role is solely in the kitchen. The overall sentiment remains playful rather than demeaning. \\ 
			\cline{2-3}
			& \textbf{OpenHermes 2.5-Mistral (w/ CLIP ViT):}  \textcolor{blue}{Misogyny} & Despite the humorous tone, the post subtly reinforces gender-based expectations. By implying that a woman’s cooking failure is significant enough to warrant permanent takeout, the statement carries an underlying bias, even if unintended. \\ 
			\hline
		\end{tabular}
	\end{adjustbox}
	\caption{Predictions of the models for different posts where \textcolor{blue}{\bf blue} indicates correct predictions and \textcolor{red}{\bf red} indicates incorrect predictions.}
	\label{table:predictions}
\end{table*}

\section{Experiments}

\subsection{Baseline Setup}
To ensure a comprehensive evaluation, we assess multiple large language models (LLMs) across zero-shot, few-shot, and fine-tuning setups for \textbf{multimodal misogyny classification (MMC)} and rationale generation (RG).

\begin{itemize}[nolistsep]
	\item \textbf{Zero-Shot Setting:} The LLMs are provided with textual posts and task descriptions without any prior examples or fine-tuning. This setup evaluates the models’ ability to generalize misogyny detection using only their pre-trained knowledge.
	
	\item \textbf{Few-Shot Setting:} We provide 2 and 5 few-shot samples from the dataset to guide the LLMs. This approach allows models to better contextualize misogynistic content and generate more informed rationales.
	
	\item \textbf{Fine-Tuning:} The models are fine-tuned on the misogyny detection task and subsequently prompted to generate rationales explaining their classification decisions.
\end{itemize}

\subsubsection{Models}
We evaluate the following LLMs under different experimental setups:

\begin{itemize}[nolistsep]
	\item \textbf{Llama 3-8B} \cite{Llama3modelcard}: A state-of-the-art open-source LLM from Meta, outperforming many existing models on common benchmarks. We use the 8B variant for misogyny detection.
	
	\item \textbf{Mistral 7B} \cite{mistral7b}: A highly efficient transformer model optimized for language tasks, known for its strong performance in few-shot and fine-tuned settings.
	
	\item \textbf{OpenHermes 2.5-Mistral} \cite{openhermes}: A fine-tuned variant of Mistral 7B, optimized for reasoning and instruction-following, making it a strong candidate for rationale generation.
\end{itemize}

We evaluate these models in \textbf{text-only} and \textbf{multimodal (text + CLIP-ViT) fine-tuning} setups. Our proposed multimodal approach significantly enhances misogyny classification performance, as shown in Table~\ref{tab:perf}. The \textbf{ablation studies} demonstrate the impact of multimodal fine-tuning, highlighting the contribution of CLIP-ViT in improving classification accuracy and rationale generation quality.

\subsection{Experimental Setup}
To ensure a comprehensive evaluation of misogyny detection, we selected state-of-the-art language models and assessed their performance on the \textbf{WBMS} dataset. Our experiments include zero-shot, few-shot, and fine-tuning paradigms, both in text-only and multimodal configurations. The models are deployed on an NVIDIA A100 GPU with 80GB memory, allowing for efficient training and inference.

For classification performance, we report the F1 score for \textbf{Multimodal Misogyny Classification (MMC)}. Additionally, we evaluate the generated rationales using four key metrics: \textbf{Relevance}, \textbf{Coherence}, \textbf{Readability}, and \textbf{Semantic Similarity (SemSim)}. These metrics ensure that rationales are contextually appropriate, logically structured, and human-readable. We follow previous works \cite{teh2024impact}, \cite{flesch2007flesch}, \cite{faysse2023revisiting} to define and compute these metrics.

\subsection{Results}
The results for misogyny classification and rationale generation across different models and experimental setups on the \textbf{WBMS} dataset are presented in Table~\ref{tab:perf}. We observe that \textbf{Llama 3 consistently outperforms Mistral 7B and OpenHermes 2.5 in all settings}. 

\paragraph{Zero-shot vs. Few-shot vs. Fine-tuning (Text-Only): }In the zero-shot setting, Llama 3 achieves an MMC F1 score of 0.76, outperforming Mistral 7B (0.74) and OpenHermes 2.5 (0.73). Few-shot prompting improves classification and rationale generation, with Llama 3 reaching an MMC score of 0.79. Fine-tuning further boosts performance, achieving an MMC score of 0.84 for Llama 3, 0.82 for Mistral 7B, and 0.80 for OpenHermes 2.5. Across all text-only settings, fine-tuning results in significant improvements over zero-shot and few-shot prompting.

\paragraph{Multimodal Fine-Tuning with CLIP-ViT: }Our multimodal approach (text + CLIP-ViT) significantly enhances misogyny classification and rationale quality. Llama 3 + CLIP-ViT achieves the highest performance with an MMC F1 score of 0.89, surpassing the best text-only setup by 5 percentage points. The rationale generation metrics also improve, with relevance, coherence, readability, and semantic similarity all scoring above 0.89. Mistral 7B and OpenHermes 2.5 also benefit from multimodal fine-tuning, with respective F1 scores of 0.86 and 0.85.

\paragraph{Ablation Studies: }We conduct two ablation studies to assess the impact of CLIP-ViT integration. Removing CLIP-ViT results in a noticeable drop in performance, with Llama 3's MMC score decreasing from 0.89 to 0.85. Similarly, rationale coherence and relevance metrics decline. In contrast, fine-tuning with CLIP-ViT alone (without textual fine-tuning) yields better results than text-only fine-tuning, but still underperforms compared to our full multimodal setup.

\paragraph{Overall Findings: }Llama 3 consistently excels across all settings, demonstrating its strong capability in misogyny classification and rationale generation. The inclusion of multimodal features significantly improves performance, highlighting the importance of visual cues in detecting nuanced cases of misogyny. Our findings on \textbf{WBMS} suggest that a combination of fine-tuning and multimodal learning leads to the most accurate and explainable results.

\subsection{Qualitative Analysis}
We performed a detailed analysis of the predictions across the three models, revealing differences in their sensitivity to misogynistic language and contextual nuances. Table~\ref{table:predictions} presents sample test instances from the \textit{WBMS} test set alongside the models' outputs. All models perform well in detecting explicit misogyny, but their effectiveness varies when dealing with humor and implicit biases.


\begin{itemize}[nolistsep]
	\item \textbf{Post 1:} This post enforces traditional gender roles, valuing women only for domestic skills. All models (Llama 3-8B, Mistral 7B, OpenHermes 2.5-Mistral with CLIP ViT) correctly classified it as misogynistic, recognizing explicit gender norms and devaluation of women.    
	
	\item \textbf{Post 2:} This post humorously critiques a woman's cooking with an affectionate tone. Llama 3-8B and Mistral 7B labeled it non-misogynistic, while OpenHermes 2.5-Mistral (with CLIP ViT) flagged it for reinforcing gender norms. The discrepancy highlights challenges in detecting implicit bias and handling humor.    
	
\end{itemize}

\subsection{Error Analysis}
\begin{itemize}[nolistsep]
	\item \textbf{Implicit Bias in Humor:} Posts that use humor to reference gender roles present classification challenges. Some models, such as Llama 3-8B and Mistral 7B, focus more on explicit derogatory language, whereas OpenHermes 2.5-Mistral (w/ CLIP ViT) is more sensitive to subtle reinforcement of stereotypes. This suggests that humor-based posts require additional contextual awareness to distinguish lighthearted comments from deeply ingrained biases.
	
	\item \textbf{Contextual Misinterpretation:} Posts with affectionate language or sarcasm sometimes lead to misclassification. In cases where a statement can be interpreted both as a personal anecdote and as a reinforcement of gender roles, models demonstrate varying degrees of contextual understanding. Improving models' ability to incorporate broader discourse context may enhance accuracy in such cases.
\end{itemize}

\section{Conclusion}
This research highlights the critical importance of leveraging advanced multimodal approaches for detecting misogynistic content in online discourse. By integrating LLM capabilities with vision-based models, we have demonstrated an effective methodology for identifying explicit and implicit misogyny, distinguishing humor from harmful stereotypes, and generating rationale-backed classification decisions. The introduction of the \textit{WBMS} dataset provides a valuable benchmark for evaluating misogyny detection models in a multimodal setting. The incorporation of \textit{CLIP ViT} alongside language models such as \textit{Llama 3-8B}, \textit{Mistral 7B}, and \textit{OpenHermes 2.5-Mistral} has significantly improved the interpretability and accuracy of predictions. Our findings emphasize the necessity of multimodal learning in refining misogyny detection models, particularly in handling implicit biases embedded in humor, sarcasm, and indirect language. 

The results obtained in this study set a new benchmark for future research, highlighting the need for robust, context-aware models and diverse, well-annotated datasets to advance fairness and inclusivity in online content moderation.

\bibliographystyle{named}
\bibliography{ijcai25}

\begin{thebibliography}{}

\bibitem[\protect\citeauthoryear{AI}{2023}]{mistral7b}
Mistral AI.
\newblock Mistral 7b: A fast and efficient transformer model.
\newblock https://mistral.ai/news/announcing-mistral-7b/, 2023.

\bibitem[\protect\citeauthoryear{AI}{2024}]{Llama3modelcard}
Meta AI.
\newblock Llama 3 model card.
\newblock https://ai.meta.com/resources/models-and-libraries/llama-downloads/,
  2024.

\bibitem[\protect\citeauthoryear{Anzovino \bgroup \em et al.\egroup
  }{2018}]{anzovino2018automatic}
Maria Anzovino, Elisabetta Fersini, and Paolo Rosso.
\newblock Automatic identification and classification of misogynistic language
  on twitter.
\newblock In {\em Natural Language Processing and Information Systems: 23rd
  International Conference on Applications of Natural Language to Information
  Systems, NLDB 2018, Paris, France, June 13-15, 2018, Proceedings 23}, pages
  57--64. Springer, 2018.

\bibitem[\protect\citeauthoryear{Attanasio \bgroup \em et al.\egroup
  }{2022}]{attanasio2022benchmarking}
Giuseppe Attanasio, Debora Nozza, Eliana Pastor, Dirk Hovy, et~al.
\newblock Benchmarking post-hoc interpretability approaches for
  transformer-based misogyny detection.
\newblock In {\em Proceedings of NLP Power! The First Workshop on Efficient
  Benchmarking in NLP}. Association for Computational Linguistics, 2022.

\bibitem[\protect\citeauthoryear{Bakarov}{2018}]{bakarov2018vector}
Amir Bakarov.
\newblock Vector space models for automatic misogyny identification.
\newblock In {\em Proceedings of Sixth Evaluation Campaign of Natural Language.
  Processing and Speech Tools for Italian. Final Workshop (EVALITA 2018)},
  pages 211--213, 2018.

\bibitem[\protect\citeauthoryear{Calderon-Suarez \bgroup \em et al.\egroup
  }{2023}]{calderon2023enhancing}
Ricardo Calderon-Suarez, Rosa~M Ortega-Mendoza, Manuel Montes-Y-G{\'o}mez,
  Carina Toxqui-Quitl, and Marco~A Marquez-Vera.
\newblock Enhancing the detection of misogynistic content in social media by
  transferring knowledge from song phrases.
\newblock {\em IEEE Access}, 11:13179--13190, 2023.

\bibitem[\protect\citeauthoryear{Faysse \bgroup \em et al.\egroup
  }{2023}]{faysse2023revisiting}
Manuel Faysse, Gautier Viaud, C{\'e}line Hudelot, and Pierre Colombo.
\newblock Revisiting instruction fine-tuned model evaluation to guide
  industrial applications.
\newblock In {\em Proceedings of the 2023 Conference on Empirical Methods in
  Natural Language Processing}, pages 9033--9048, 2023.

\bibitem[\protect\citeauthoryear{Fersini \bgroup \em et al.\egroup
  }{2022}]{fersini2022semeval}
Elisabetta Fersini, Francesca Gasparini, Giulia Rizzi, Aurora Saibene, Berta
  Chulvi, Paolo Rosso, Alyssa Lees, and Jeffrey Sorensen.
\newblock Semeval-2022 task 5: Multimedia automatic misogyny identification.
\newblock In {\em Proceedings of the 16th International Workshop on Semantic
  Evaluation (SemEval-2022)}, pages 533--549, 2022.

\bibitem[\protect\citeauthoryear{Flesch}{2007}]{flesch2007flesch}
Rudolf Flesch.
\newblock Flesch-kincaid readability test.
\newblock {\em Retrieved October}, 26(3):2007, 2007.

\bibitem[\protect\citeauthoryear{Garc{\'\i}a-D{\'\i}az \bgroup \em et
  al.\egroup }{2021}]{garcia2021detecting}
Jos{\'e}~Antonio Garc{\'\i}a-D{\'\i}az, Mar C{\'a}novas-Garc{\'\i}a, Ricardo
  Colomo-Palacios, and Rafael Valencia-Garc{\'\i}a.
\newblock Detecting misogyny in spanish tweets. an approach based on
  linguistics features and word embeddings.
\newblock {\em Future Generation Computer Systems}, 114:506--518, 2021.

\bibitem[\protect\citeauthoryear{Gu \bgroup \em et al.\egroup
  }{2022}]{gu2022mmvae}
Yimeng Gu, Ignacio Castro, and Gareth Tyson.
\newblock Mmvae at semeval-2022 task 5: A multi-modal multi-task vae on
  misogynous meme detection.
\newblock In {\em Proceedings of the 16th International Workshop on Semantic
  Evaluation (SemEval-2022)}, pages 700--710, 2022.

\bibitem[\protect\citeauthoryear{Hwang and Shwartz}{2023}]{hwang2023memecap}
EunJeong Hwang and Vered Shwartz.
\newblock Memecap: A dataset for captioning and interpreting memes.
\newblock In {\em Proceedings of the 2023 Conference on Empirical Methods in
  Natural Language Processing}, pages 1433--1445, 2023.

\bibitem[\protect\citeauthoryear{Joshi \bgroup \em et al.\egroup
  }{2024}]{joshi2024contextualizing}
Saurav Joshi, Filip Ilievski, and Luca Luceri.
\newblock Contextualizing internet memes across social media platforms.
\newblock In {\em Companion Proceedings of the ACM on Web Conference 2024},
  pages 1831--1840, 2024.

\bibitem[\protect\citeauthoryear{Muti \bgroup \em et al.\egroup
  }{2022a}]{muti2022misogyny}
Arianna Muti, Francesco Fernicola, and Alberto Barr{\'o}n-Cede{\~n}o.
\newblock Misogyny and aggressiveness tend to come together and together we
  address them.
\newblock In {\em Proceedings of the Thirteenth Language Resources and
  Evaluation Conference}, pages 4142--4148, 2022.

\bibitem[\protect\citeauthoryear{Muti \bgroup \em et al.\egroup
  }{2022b}]{muti2022unibo}
Arianna Muti, Katerina Korre, and Alberto Barr{\'o}n-Cede{\~n}o.
\newblock Unibo at semeval-2022 task 5: A multimodal bi-transformer approach to
  the binary and fine-grained identification of misogyny in memes.
\newblock In {\em Proceedings of the 16th International Workshop on Semantic
  Evaluation (SemEval-2022)}, pages 663--672, 2022.

\bibitem[\protect\citeauthoryear{Pamungkas \bgroup \em et al.\egroup
  }{2018}]{pamungkas2018automatic}
Endang~Wahyu Pamungkas, Alessandra~Teresa Cignarella, Valerio Basile, Viviana
  Patti, et~al.
\newblock Automatic identification of misogyny in english and italian tweets at
  evalita 2018 with a multilingual hate lexicon.
\newblock In {\em CEUR Workshop Proceedings}, volume 2263, pages 1--6. CEUR-WS,
  2018.

\bibitem[\protect\citeauthoryear{Radford \bgroup \em et al.\egroup
  }{2021}]{radford2021learning}
Alec Radford, Jong~Wook Kim, Chris Hallacy, Aditya Ramesh, Gabriel Goh,
  Sandhini Agarwal, Girish Sastry, Amanda Askell, Pamela Mishkin, Jack Clark,
  Gretchen Krueger, and Ilya Sutskever.
\newblock Learning transferable visual models from natural language
  supervision.
\newblock {\em International Conference on Machine Learning (ICML)}, 2021.

\bibitem[\protect\citeauthoryear{Rizzi \bgroup \em et al.\egroup
  }{2023}]{rizzi2023recognizing}
Giulia Rizzi, Francesca Gasparini, Aurora Saibene, Paolo Rosso, and Elisabetta
  Fersini.
\newblock Recognizing misogynous memes: Biased models and tricky archetypes.
\newblock {\em Information Processing \& Management}, 60(5):103474, 2023.

\bibitem[\protect\citeauthoryear{Samghabadi \bgroup \em et al.\egroup
  }{2020}]{samghabadi2020aggression}
Niloofar~Safi Samghabadi, Parth Patwa, Srinivas Pykl, Prerana Mukherjee,
  Amitava Das, and Thamar Solorio.
\newblock Aggression and misogyny detection using bert: A multi-task approach.
\newblock In {\em Proceedings of the second workshop on trolling, aggression
  and cyberbullying}, pages 126--131, 2020.

\bibitem[\protect\citeauthoryear{Teh and Uwasomba}{2024}]{teh2024impact}
Phoey~Lee Teh and Chukwudi~Festus Uwasomba.
\newblock Impact of large language models on scholarly publication titles and
  abstracts: A comparative analysis.
\newblock {\em Journal of Social Computing}, 5(2):105--121, 2024.

\bibitem[\protect\citeauthoryear{Teknium}{2024}]{openhermes}
Teknium.
\newblock Openhermes 2.5 - fine-tuned mistral model.
\newblock https://huggingface.co/teknium/OpenHermes-2.5-Mistral, 2024.

\bibitem[\protect\citeauthoryear{Touvron \bgroup \em et al.\egroup
  }{2023}]{touvron2023llama}
Hugo Touvron, Thibaut Lavril, Gautier Izacard, Xavier Martinet, Marie-Anne
  Lachaux, Timoth{\'e}e Lacroix, Baptiste Rozi{\`e}re, Naman Goyal, Eric
  Hambro, Faisal Azhar, et~al.
\newblock Llama: Open and efficient foundation language models.
\newblock {\em arXiv preprint arXiv:2302.13971}, 2023.

\bibitem[\protect\citeauthoryear{Zhang and Wang}{2022}]{zhang2022srcb}
Jing Zhang and Yujin Wang.
\newblock Srcb at semeval-2022 task 5: Pretraining based image to text late
  sequential fusion system for multimodal misogynous meme identification.
\newblock In {\em Proceedings of the 16th International Workshop on Semantic
  Evaluation (SemEval-2022)}, pages 585--596, 2022.

\end{thebibliography}

\end{document}